\documentclass[letterpaper, 10 pt, conference]{ieeeconf}
\IEEEoverridecommandlockouts    
\usepackage{graphics}           
\usepackage{times}              
\usepackage{amsmath}            
\usepackage{amssymb}            
\usepackage{graphicx}
\usepackage{algorithm}
\usepackage[noend]{algpseudocode}
\usepackage{booktabs}
\usepackage{color}
\usepackage[nocompress]{cite} 
\definecolor{instructioncolor}{rgb}{.5,.5,.5}

\usepackage[font=small]{caption}

\def\eqref#1{Eq.~(\ref{#1})}

\makeatletter
\usepackage{xspace}
\DeclareRobustCommand\onedot{\futurelet\@let@token\@onedot}
\def\@onedot{\ifx\@let@token.\else.\null\fi\xspace}

\makeatother

\usepackage{array}
\newcolumntype{L}[1]{>{\raggedright\let\newline\\\arraybackslash\hspace{0pt}}m{#1}}
\newcolumntype{C}[1]{>{\centering\let\newline\\\arraybackslash\hspace{0pt}}m{#1}}
\newcolumntype{R}[1]{>{\raggedleft\let\newline\\\arraybackslash\hspace{0pt}}m{#1}}

\usepackage{flushend}
\usepackage{dsfont}
\usepackage{graphicx}
\usepackage[colorlinks=true,
            citecolor=blue,
            linkcolor=blue,
            anchorcolor=blue,
            urlcolor=blue]{hyperref}
\usepackage{amsmath}
\usepackage{amssymb}
\usepackage{tabularx}
\usepackage{bbm}
\usepackage{multirow}
\usepackage{cite}
\usepackage{paralist}

\makeatletter
\renewcommand{\maketag@@@}[1]{\hbox{\m@th\small\normalfont#1}}%
\makeatother

\newcommand{\keywordss}[1]{\par\textbf{\textit{Index Terms---}}\textbf{#1}}

\title{\LARGE \bf TOL: Textual Localization with OpenStreetMap}

\author{
Youqi Liao,
Shuhao Kang, 
Jingyu Xu,  
Olaf Wysocki, 
Yan Xia,
Jianping Li\textsuperscript{\textdagger}, 
Zhen Dong,
Bisheng Yang,
Xieyuanli Chen
\thanks{
This research was supported by the National Natural Science Foundation Project (No.42201477, No. 42130105). (Corresponding author: Jianping Li)}
\thanks{
 Y. Liao, Z. Dong and B. Yang are with Wuhan University. S. Kang is with the Technical University of Munich. J. Li is with Nanyang Technological University. J. Xu is with the Institute of Artificial Intelligence (TeleAI), China Telecom, China. Olaf Wysocki is with the University of Cambridge, CV4DT, UK. Y. Xia is with the  University of Science and Technology of China, China. X. Chen is with the national university of defense technology, China.}
}

\begin{document}
\maketitle
\markboth{IEEE Robotics and Automation Letters}%
{Y. Liao, \MakeLowercase{\textit{et al.}}: TOL: Textual Localization with OpenStreetMap}

\setlength{\textfloatsep}{1.3em}
\setlength{\dbltextfloatsep}{1.3em}

\begin{abstract}
Natural language provides an intuitive way to express spatial intent in geospatial applications. While existing localization methods often rely on dense point cloud maps or high-resolution imagery, OpenStreetMap (OSM) offers a compact and freely available map representation that encodes rich semantic and structural information, making it well-suited for large-scale localization. However, text-to-OSM (T2O) localization remains largely unexplored.
In this paper, we formulate the T2O localization task, which aims to estimate accurate 2D positions in urban environments from textual scene descriptions without relying on geometric observations or GNSS-based initial location. To support the proposed task, we introduce TOL, a large-scale benchmark spanning multiple continents and diverse urban environments. TOL contains approximately 121K textual queries paired with OSM map tiles and covers about 316 km of road trajectories across Boston, Karlsruhe, and Singapore.
We further propose TOLoc, a coarse-to-fine localization framework that explicitly models the semantics of surrounding objects and their directional information. In the coarse stage, direction-aware features are extracted from both textual descriptions and OSM tiles to construct global descriptors, which are used to retrieve candidate locations for the query. In the fine stage, the query text and top-1 retrieved tile are jointly processed, where a dedicated alignment module fuses the textual descriptor and local map features to regress the 2-DoF pose. Experimental results demonstrate that TOLoc achieves strong localization performance, outperforming the best existing method by 6.53\%, 9.93\%, and 8.32\% at 5 m, 10 m, and 25 m thresholds, respectively, and shows strong generalization to unseen environments. Dataset, code and models will be publicly available
at: \url{https://github.com/WHU-USI3DV/TOL}.
\end{abstract}

\keywordss{ Textual localization, OpenStreetMap, Urban Scene Understanding}

\section{Introduction}

\begin{figure}[t]
\centering
\includegraphics[width=3.2in]{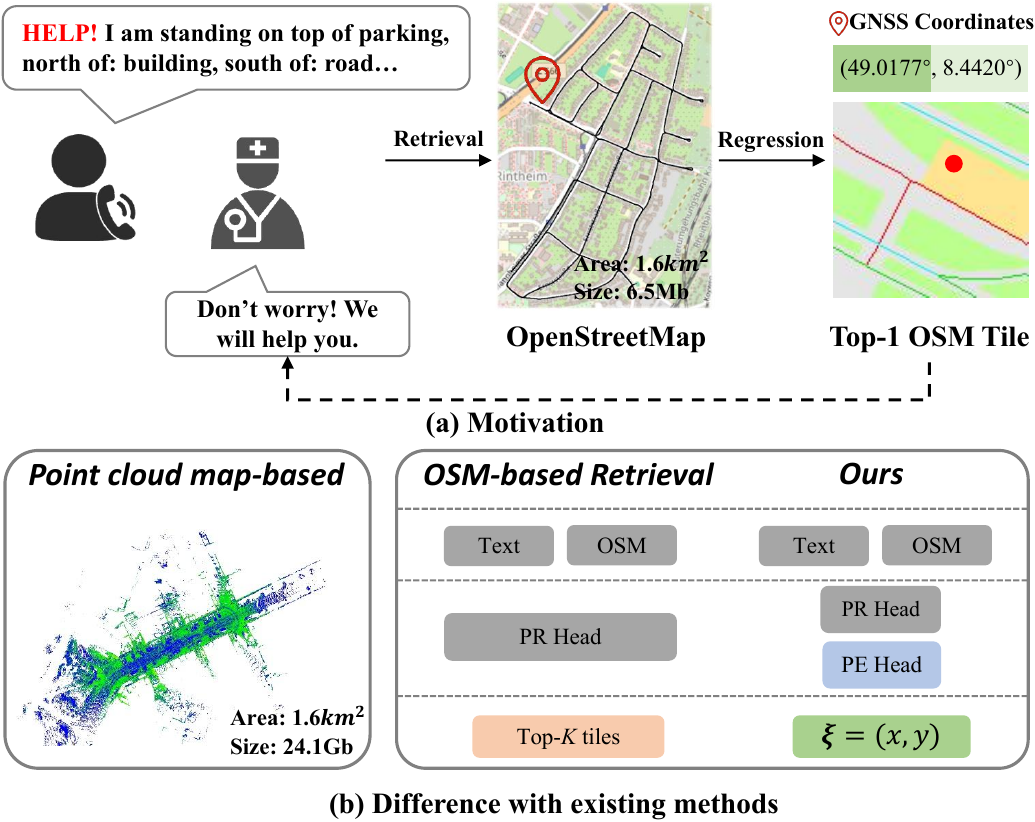}
\vspace{-4pt}
\caption{Overview and motivation of text-to-OSM localization. (a) Given a textual query, the system retrieves the most relevant OSM tile and estimates the final 2-DoF position. (b) Compared with point cloud map-based localization, OSM provides a lightweight map representation. Unlike existing text-driven retrieval methods that only return candidate tiles, our method further performs pose estimation for meter-level localization.}
\vspace{-6pt}
\label{fig:motivation}
\end{figure}

Text-based descriptions provide a natural way to communicate spatial cues in real-world environments~\cite{giudice2007wayfinding,li2025cityanchor}. Accordingly, text-driven localization aims to infer a target location by aligning the semantic information in a description with a map representation~\cite{xia2024text2loc,kang2026vlm}. This capability is particularly meaningful in human--robot interaction. For example, when an elderly person becomes lost or encounters difficulties in an unfamiliar environment, the surrounding scene is often described in natural language, as shown in Fig.~\ref{fig:motivation}(a). A robot or rescue system can then interpret the description and match it against a map to estimate the person's location and provide timely assistance. Such a capability is valuable for applications such as emergency response and assistive urban navigation, where reliable localization from human instructions is essential.

To date, most localization methods rely on image queries~\cite{hausler2021patch,yin2023isimloc} or point cloud queries~\cite{luo2025bevplace++,kang2025opal}, while text-driven localization remains less explored. Early studies mainly focus on text-to-point-cloud (T2P) localization, pioneered by Text2Pos~\cite{kolmet2022text2pos} and extended by subsequent methods~\cite{wang2023text,xia2024text2loc,xu2025cmmloc}. More recently, GOTPR~\cite{jung2025gotpr} and CVG-Text~\cite{ye2025cross} have investigated text-to-OSM (T2O) place recognition. GOTPR represents textual descriptions and OSM maps as scene graphs, and performs retrieval through graph matching. CVG-Text constructs text--OSM pairs from geo-tagged image--OSM correspondences, using semantic extraction and LLM-based text generation~\cite{hurst2024gpt}. It then studies cross-modal retrieval for both T2A and T2O settings. As illustrated in Fig.~\ref{fig:motivation}(b), existing T2O methods mainly perform OSM-based retrieval, whereas practical localization requires further pose estimation within the retrieved map tile. In addition, GOTPR relies on graph matching, which becomes less scalable in complex scenes, while CVG-Text requires LLM-based generation and manual post-processing, increasing data preparation cost. These limitations motivate a scalable T2O localization framework that supports both place recognition and accurate pose estimation.

To address these limitations, we study the T2O localization task, which aims to estimate the 2-DoF position of a textual query in city-scale OSM maps. To support this task, we introduce \textbf{TOL}, a large-scale benchmark containing about 121K text--OSM pairs from diverse urban scenes in Singapore, Boston, and Karlsruhe. The textual descriptions are automatically constructed from visible OSM objects around the query position using predefined directional templates, resulting in a fully automated and scene-agnostic pipeline without manual annotation or LLM-based text generation.
To solve the proposed task, we present \textbf{TOLoc}, a two-stage framework that first performs place recognition to retrieve candidate OSM tiles and then refines the query position through pose estimation. Specifically, we build direction-aware text and map representations by aggregating semantic features from different directions, enabling effective cross-modal retrieval between textual queries and OSM tiles. Given the top-ranked tile, we further introduce a text-to-OSM alignment (TOA) module to fuse the textual descriptor with the local OSM feature map for accurate 2-DoF pose prediction. Experiments show that TOLoc achieves over 28\% localization success within a 25-meter threshold, outperforming the state-of-the-art method CVG-Text~\cite{ye2025cross} by 9.87\%. Additional cross-scene evaluations further demonstrate strong generalization to geographically distant and previously unseen environments. Our contributions are summarized as follows:
\begin{itemize}
    \item We introduce T2O localization, a new task that estimates the 2-DoF position of a textual query in city-scale OSM maps, and present TOL, a large-scale benchmark for systematic evaluation in diverse urban scenes.
    \item We propose TOLoc, a two-stage framework for T2O localization.  By fully exploiting direction-aware semantic cues from both map and textual modalities, TOLoc constructs robust and discriminative representations for localization.
    \item Experiments show that TOLoc outperforms existing methods in place recognition, localization accuracy, and cross-scene generalization.
\end{itemize}

The remainder of this paper is organized as follows. Sec.~\ref{sec:Related Works} reviews related work on text-driven and OSM-based localization. Sec.~\ref{sec:TOLoc} introduces the TOL benchmark and its construction process. Sec.~\ref{sec:method} presents the proposed TOLoc framework. Sec.~\ref{sec:exp} reports the experimental results, ablation studies, and failure case analysis. Sec.~\ref{sec:conclusion} concludes the paper.

\section{Related Work}\label{sec:Related Works}

\subsection{Text-driven Localization}
Text-driven localization estimates spatial positions from natural language descriptions. One major direction is T2P localization. Text2Pos~\cite{kolmet2022text2pos} first formulates this task with a two-stage pipeline of retrieval followed by pose estimation. Subsequent methods improve text--point cloud alignment through transformer-based encoders~\cite{wang2023text,xia2024text2loc}, multi-level contrastive learning~\cite{liu2025text}, probabilistic priors with directional cues~\cite{xu2025cmmloc}, uncertainty-aware partial matching~\cite{feng2025partially}, and VLM-based reasoning~\cite{kang2026vlm}. Despite promising results, these methods rely on large-scale point cloud maps, which are costly to build, store, and maintain, limiting their practicality in dynamic real-world environments.

Another line of work studies text-driven geo-localization with aerial or multi-view imagery. GeoText-1652~\cite{chu2024towards} extends University-1652~\cite{zheng2020university} with spatially aware descriptions for T2A localization, and HCCM~\cite{ruan2025hccm} further improves cross-modal matching with hierarchical multi-scale learning. CVG-Text~\cite{ye2025cross} broadens this setting by incorporating panoramic images and OSM tiles, while MMGeo~\cite{ji2025mmgeo} introduces additional modalities such as depth maps and point clouds. Although aerial and UAV imagery is easier to acquire than point cloud maps, it still requires substantial storage and processing, and remains sensitive to seasonal, illumination, and environmental changes. In contrast, OSM is lightweight, freely available, and continuously updated, making it attractive for scalable text-based localization. However, localization directly on OSM maps remains underexplored, especially beyond place recognition toward accurate pose estimation.

\subsection{OSM-based Localization}

Most OSM-based localization methods use images or point clouds as queries. For image-to-OSM (I2O) localization, early methods integrate OSM-derived road constraints into visual odometry or probabilistic localization frameworks~\cite{floros2013openstreetslam,zhou2021efficient}, while retrieval-based approaches match query images with OSM patches~\cite{samano2020you}. More recent works focus on single-image I2O localization. OrienterNet~\cite{sarlin2023orienternet} introduces an end-to-end neural matching framework, MapLocNet~\cite{wu2024maplocnet} adopts a two-stage pose regression pipeline, and OSMLoc~\cite{liao2024osmloc} further exploits geometric and semantic guidance. Although effective, I2O methods rely on visual observations, which may be affected by illumination, viewpoint, and appearance changes.

Point cloud-to-OSM (P2O) localization has also been widely studied. Early methods align road or building features from sequential point clouds with OSM data~\cite{ruchti2015localization,vysotska2016exploiting}, while later approaches incorporate semantic cues into probabilistic localization or LiDAR--inertial systems~\cite{suger2017global,lee2024autonomous}. For single-scan localization, hand-crafted descriptors~\cite{cho2022openstreetmap} and learning-based methods such as OPAL~\cite{kang2025opal} match LiDAR scans with OSM tiles. However, P2O methods require dense geometric observations, which are difficult for humans to provide in language-based human--robot interaction.

Compared with images and point clouds, textual descriptions offer a natural way to express high-level semantics and spatial relations. Nevertheless, T2O localization remains underexplored. CVG-Text~\cite{ye2025cross} and GOTPR~\cite{jung2025gotpr} are the only prior works that explicitly study T2O place recognition: the former performs cross-modal retrieval between text and OSM tiles, while the latter matches scene graphs built from both modalities. In contrast, we study T2O  localization beyond tile retrieval, aiming at meter-level pose estimation.

\section{TOL benchmark}\label{sec:TOLoc}

\subsection{Task Definition}
In this paper, we study T2O localization, where the goal is to estimate the 2-DoF position $\boldsymbol{\xi}=(x,y)$ of a query from its textual description $\mathcal{T}$ in a city-scale OSM map. The description $\mathcal{T}=\{h_i\}_{i=1}^{N}$ consists of multiple linguistic hints that describe the semantics of surrounding objects and their relative directions with respect to $\boldsymbol{\xi}$. We formulate this task in a coarse-to-fine manner, including a place recognition (PR) stage that retrieves the top-$K$ candidate OSM tiles from the database and a pose estimation (PE) stage that predicts the final 2-DoF position within the top-1 retrieved tile.

\subsection{TOL Benchmark}
Existing datasets and resources~\cite{jung2025gotpr,ye2025cross} are insufficient for T2O localization, as they lack text--OSM pairs with meter-level position annotations. To fill this gap, we introduce \textbf{TOL}, a large-scale benchmark built from publicly available OSM data\footnote{\url{https://www.openstreetmap.org/}} and vehicle trajectories from NuScenes~\cite{caesar2020nuscenes} and KITTI-360~\cite{liao2022kitti}. TOL covers Singapore, Boston, and Karlsruhe, and provides spatially grounded text--OSM pairs generated through a fully automated, scene-agnostic pipeline without manual annotation or LLM-based text generation.

\subsubsection{Map Construction}
\begin{figure}[t]
\centering
\includegraphics[width=3.3in]{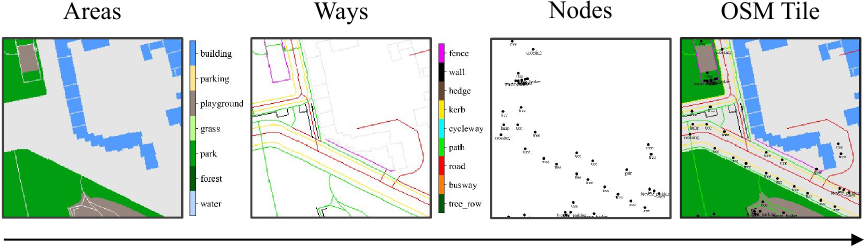}
\vspace{-3pt}
\caption{Illustration of rasterized OSM tiles with area, way, and node channels.}
\label{fig_osm_tile}
\end{figure}

We construct the map database from publicly available OSM data along the GNSS trajectories of NuScenes~\cite{caesar2020nuscenes} and KITTI-360~\cite{liao2022kitti}. For each frame, we crop an $H \times H$ meter OSM tile centered at the GNSS location, project the selected map elements into a local Cartesian coordinate system, and render the elements listed in Tab.~\ref{tab:osm} within the tile. Following prior work~\cite{sarlin2023orienternet,kang2025opal}, we group OSM elements into node, way, and area channels, and rasterize each tile as a 3-channel grid image $\mathcal{O} \in \mathbb{R}^{\frac{H}{\Delta} \times \frac{H}{\Delta} \times 3}$ with sampling distance $\Delta$, as shown in Fig.~\ref{fig_osm_tile}. These tiles form a city-scale OSM database covering the full vehicle trajectories. We set $H=50$ m and $\Delta=0.25$ m/pixel.

\begin{table}[t]
\caption{Details of OSM elements.}
\label{tab:osm}
\centering
\footnotesize
\setlength{\tabcolsep}{4pt}
\renewcommand{\arraystretch}{1.1}
\begin{tabularx}{\columnwidth}{p{1.1cm}X}
\hline
\textbf{Type} & \textbf{Element} \\ \hline
Areas & building, parking, playground, grass, park, forest, water \\ \hline
Ways & fence, wall, hedge, kerb, cycleway, path, road, busway, tree row \\ \hline
Nodes & parking entrance, street lamp, junction, traffic signal, stop sign, give way sign, bus stop, stop area, crossing, gate, bollard, gas station, bicycle parking, charging station, shop, restaurant, bar, vending machine, pharmacy, tree, stone, ATM, toilets, water fountain, bench, waste basket, post box, artwork, recycling station, clock, fire hydrant, pole, street cabinet \\ \hline
\end{tabularx}
\vspace{-7pt}
\end{table}

\subsubsection{Text Query Construction}
For each OSM tile $\mathcal{O}$, we sample a text-query position within a square region of side length $\frac{H}{2}$ centered at the tile center. Since OSM data is sparse, we preferentially sample from cells containing valid semantic elements; otherwise, we uniformly sample a cell from the region. This strategy increases the discriminability of generated queries.

Given the query position, we construct a five-sentence description $\mathcal{T}=\{T_t,T_n,T_s,T_w,T_e\}$, where $T_t$ describes the semantic cue around the query position, and $T_n,T_s,T_w,T_e$ describe visible objects in the north, south, west, and east directions, respectively. Candidate objects are collected within a circular region of radius $\frac{H}{2}$ around the query position.

To model visibility, we follow OPAL~\cite{kang2025opal} and treat \texttt{building} elements as dominant occluders. The circular region is discretized into a polar grid with $U$ radial bins and $V$ angular sectors, with resolutions $\Delta_r=\frac{H}{2U}$ and $\Delta_a=\frac{2\pi}{V}$. Each cell $(u,v)$ has radial distance and azimuth angle:
\begin{equation}
\begin{aligned}
r_{u,v}&=(u+0.5)\Delta_r, \quad u\in\{0,\dots,U-1\},\\
\phi_{u,v}&=(v+0.5)\Delta_a, \quad v\in\{0,\dots,V-1\}.
\end{aligned}
\label{eq:polar_grid}
\end{equation}
Ray casting is then performed along each angular sector to obtain the visibility mask:
\begin{equation}
\mathcal{M}(u,v)=
\begin{cases}
\text{False}, & u>\min(u') \ \text{if} \ \exists \mathcal{O}(u',v)=\text{``building"},\\
\text{True}, & \text{otherwise},
\end{cases}
\label{eq:visibility_mask}
\end{equation}
from which the visible object set is defined as
$\mathcal{X}_{\text{vis}}=\{x(u,v)\mid \mathcal{M}(u,v)=\text{True}\}$.

Visible objects are grouped into one center region and four cardinal directions based on their relative positions to the query location. For an object $x\in\mathcal{X}_{\text{vis}}$ in polar cell $(u,v)$, its directional label is defined as
\begin{equation}
\text{direction}(u,v)=
\begin{cases}
\text{top}, & r_{u,v}\leq \sigma,\\
\text{east}, & r_{u,v}>\sigma,\ \phi_{u,v}<\frac{\pi}{4} \ \text{or} \ \phi_{u,v}\geq\frac{7\pi}{4},\\
\text{north}, & r_{u,v}>\sigma,\ \frac{\pi}{4}\leq\phi_{u,v}<\frac{3\pi}{4},\\
\text{west}, & r_{u,v}>\sigma,\ \frac{3\pi}{4}\leq\phi_{u,v}<\frac{5\pi}{4},\\
\text{south}, & r_{u,v}>\sigma,\ \frac{5\pi}{4}\leq\phi_{u,v}<\frac{7\pi}{4},
\end{cases}
\label{eq:direction}
\end{equation}
where $\sigma$ controls the center-region size.

The grouped cues are then converted into text using predefined templates: \texttt{"The pose is on top of <semantic>."} for the center region and \texttt{"The pose is <direction> of <semantic>."} for each cardinal direction. If no valid object is observed, \texttt{<semantic>} is set to \texttt{None}. Thus, each query contains five sentences. We set $U=\frac{H}{2}$, $V=360$, and $\sigma=3$.

\subsubsection{Data Statistics}
Using the above pipeline, we construct TOL across three cities and organize it into two subsets: TOL-N and TOL-K360. TOL-N contains 34,149 text--OSM pairs from four scenes in Singapore and Boston, covering approximately 242 km of road trajectories, while TOL-K360 contains 87,108 pairs from 11 Karlsruhe sequences, covering approximately 74 km. We use three TOL-N scenes for training (30,722 pairs) and the remaining scene for validation (3,427 pairs). Cross-scene generalization is evaluated on each TOL-K360 sequence.

\section{Methodology}\label{sec:method} 

\begin{figure*}[t]
\centering
\includegraphics[width=0.9\linewidth]{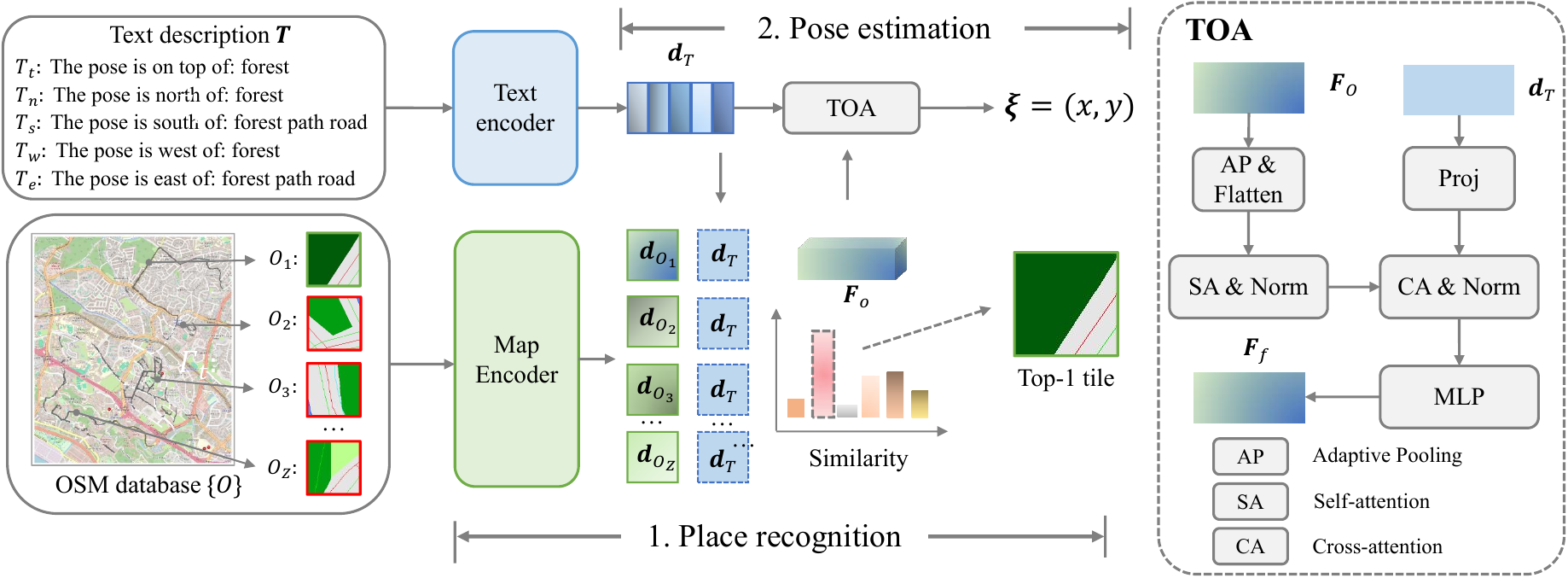}
\vspace{-4pt}
\caption{Pipeline of TOLoc. Given a query text $\mathcal{T}$ and an OSM database $\mathbb{O}=\{\mathcal{O}_j\}_{j=1}^{Z}$, TOLoc performs coarse-to-fine localization. It first learns text--map correspondences to retrieve the top-$K$ candidate tiles, and then aligns the text descriptor $\boldsymbol{d}_{\mathcal{T}}$ with the feature map $\boldsymbol{F}_{\mathcal{O}}$ of the top-1 tile to predict the final 2-DoF position. The proposed TOA module fuses local OSM features and textual cues through self-attention and cross-attention for fine-grained localization.}
\label{fig:workflow}
\vspace{-4pt}
\end{figure*}

To address the T2O localization task, we propose \textbf{TOLoc}, as illustrated in Fig.~\ref{fig:workflow}. TOLoc follows a two-stage coarse-to-fine localization pipeline. Given a city-scale OSM database $\mathbb{O}=\{\mathcal{O}_j\}_{j=1}^{Z}$, the query text $\mathcal{T}$ and OSM tiles are encoded by a dual text--map encoder to produce textual and map features. In the place recognition stage (Sec.~\ref{subsec: pr}), each OSM feature map is partitioned into five directional regions, namely \texttt{top}, \texttt{north}, \texttt{south}, \texttt{west}, and \texttt{east}. Features within each region are aggregated and concatenated to form a global map descriptor $\boldsymbol{d}_{\mathcal{O}_j}$, while the directional text features are concatenated in the same order to form the text descriptor $\boldsymbol{d}_{\mathcal{T}}$. Similarity scores between the query descriptor and all map descriptors are then computed to retrieve the top-$K$ candidate tiles. In the subsequent pose estimation stage (Sec.~\ref{subsec: pe}), the textual features and the feature map of the top-1 retrieved tile are fed into the proposed TOA module for cross-modal fusion, from which the final 2-DoF position is regressed. Sec.~\ref{subsec:training} further describes the training strategy, inference procedure, and loss functions.

\subsection{Place Recognition}\label{subsec: pr}
Goal of the PR stage is to narrow the search space by retrieving the top-$K$ OSM tiles that are most similar to the query text, which are then used as candidates for the subsequent pose estimation stage. To this end, the query text and each OSM tile are encoded into high-dimensional descriptors, and their similarities are computed for retrieval.

Given a textual description $\mathcal{T}=\{h_t,h_n,h_s,h_w,h_e\}$, we first encode each directional hint using the text encoder $\mathrm{E}_{\mathcal{T}}$:
\begin{equation}
\boldsymbol{f}_{\mathcal{T}}^r = \mathrm{E}_{\mathcal{T}}(h_r) \in \mathbb{R}^{d}, \qquad r \in \{t,n,s,w,e\}.
\end{equation}
The resulting features are concatenated and passed through a multi-layer perceptron (MLP) to obtain the text descriptor:
\begin{equation}
\boldsymbol{d}_{\mathcal{T}} = \mathrm{MLP}\big([\boldsymbol{f}_{\mathcal{T}}^t;\boldsymbol{f}_{\mathcal{T}}^n;\boldsymbol{f}_{\mathcal{T}}^s;\boldsymbol{f}_{\mathcal{T}}^w;\boldsymbol{f}_{\mathcal{T}}^e]\big) \in \mathbb{R}^{d}.
\end{equation}

For each map tile $\mathcal{O}_j$ in the database, we extract a spatial feature map $\boldsymbol{F}_{\mathcal{O}_j}$ using the map encoder $\mathrm{E}_{\mathcal{O}}$, where $\boldsymbol{F}_{\mathcal{O}_j} \in \mathbb{R}^{Q \times Q \times C}$ denotes the patch-level map representation. To construct a direction-aware map descriptor, we partition the feature map into five regions according to the polar-coordinate rule in Eq.~(\ref{eq:direction}), corresponding to \texttt{top}, \texttt{north}, \texttt{south}, \texttt{west}, and \texttt{east}. Based on this partition, we define five binary masks $\mathcal{M}=\{M^t,M^n,M^s,M^w,M^e\}$, where each mask $M^r \in \{0,1\}^{Q \times Q}$ is defined as
\begin{equation}
M^r(x,y)=
\begin{cases}
1, & \text{if } \operatorname{direction}(u,v)=r,\\
0, & \text{otherwise},
\end{cases}
\end{equation}
with $(u,v)$ denoting the polar coordinates of grid $(x,y)$.

Feature of each directional region is then obtained by masked average pooling:
\begin{equation}
\begin{aligned}
   \boldsymbol{f}_{\mathcal{O}_j}^r =
 \frac{1}{\sum_{x,y} M^r(x,y)}
& \sum_{x=1}^{Q}\sum_{y=1}^{Q}
M^r(x,y)\,\boldsymbol{F}_{\mathcal{O}_j}(x,y), \\
& \qquad r \in \{t,n,s,w,e\}, 
\end{aligned}
\end{equation}
where $\boldsymbol{f}_{\mathcal{O}_j}^r \in \mathbb{R}^{C}$ denotes the aggregated feature of the $r$-th directional region. The final map descriptor is constructed by concatenating the five directional features and projecting them with an MLP:
\begin{equation}
\boldsymbol{d}_{\mathcal{O}_j} =
\mathrm{MLP}\big([\boldsymbol{f}_{\mathcal{O}_j}^t;\boldsymbol{f}_{\mathcal{O}_j}^n;\boldsymbol{f}_{\mathcal{O}_j}^s;\boldsymbol{f}_{\mathcal{O}_j}^w;\boldsymbol{f}_{\mathcal{O}_j}^e]\big)
\in \mathbb{R}^{d}.
\end{equation}

Given the query descriptor $\boldsymbol{d}_{\mathcal{T}}$ and the map descriptors $\boldsymbol{d}_{\mathcal{O}_j}$, we compute their cosine similarities:
\begin{equation}\label{eq:cos_sim}
S_{\mathcal{T}\rightarrow\mathcal{O}_j}
=
\operatorname{sim}(\boldsymbol{d}_{\mathcal{T}}, \boldsymbol{d}_{\mathcal{O}_j}),
\qquad j=1,\dots,Z,
\end{equation}
where $\operatorname{sim}(\cdot,\cdot)$ denotes cosine similarity. The top-$K$ candidate tiles $\mathcal{K}$ are selected according to the similarity scores.

Our framework is backbone-agnostic. In practice, we instantiate the text and map encoders with CLIP~\cite{radford2021learning} and SigLIP~\cite{zhai2023sigmoid}, two widely used contrastive learning models.

\subsection{Pose Estimation}\label{subsec: pe}
In the PE stage, the goal is to predict the 2-DoF position within the top-1 retrieved OSM tile. Unlike the PR stage, which performs global retrieval over the map database, PE focuses on fine-grained local reasoning within the selected tile. To this end, we introduce the text-to-OSM alignment (TOA) module, shown in the last column of Fig.~\ref{fig:workflow}, to fuse local map features with the textual descriptor for precise position regression.

Given the patch-level map features $\boldsymbol{F}_{\mathcal{O}} \in \mathbb{R}^{Q \times Q \times C}$ of the top-1 retrieved tile and the text descriptor $\boldsymbol{d}_{\mathcal{T}} \in \mathbb{R}^{d}$, we first project them into a shared $C$-dimensional space:
\begin{equation}
\boldsymbol{F}_{\mathcal{O}}' = \mathrm{MLP}_{\mathcal{O}}(\boldsymbol{F}_{\mathcal{O}}), \qquad
\boldsymbol{d}_{\mathcal{T}}' = \mathrm{MLP}_{\mathcal{T}}(\boldsymbol{d}_{\mathcal{T}}).
\end{equation}

The projected map features $\boldsymbol{F}_{\mathcal{O}}'$ are then flattened into a $(Q^2)\times C$ matrix and processed by a self-attention layer to capture long-range spatial dependencies within the tile:
\begin{equation}
\boldsymbol{F}_{\mathcal{O}}^{\mathrm{sa}}
=
\operatorname{softmax}
\Big(
\frac{\boldsymbol{F}_{\mathcal{O}}'(\boldsymbol{F}_{\mathcal{O}}')^{\top}}{\sqrt{C}}
\Big)
\boldsymbol{F}_{\mathcal{O}}'.
\end{equation}

Next, cross-attention is applied between the refined map features $\boldsymbol{F}_{\mathcal{O}}^{\mathrm{sa}}$ and the projected text descriptor $\boldsymbol{d}_{\mathcal{T}}'$ to inject semantic and directional cues from the text into the local map representation:
\begin{equation}
\boldsymbol{F}_{f}
=
\operatorname{softmax}
\Big(
\frac{\boldsymbol{F}_{\mathcal{O}}^{\mathrm{sa}}(\boldsymbol{d}_{\mathcal{T}}')^{\top}}{\sqrt{C}}
\Big)
\boldsymbol{F}_{\mathcal{O}}^{\mathrm{sa}}.
\end{equation}

Finally, the fused representation $\boldsymbol{F}_{f} \in \mathbb{R}^{Q^2 \times C}$ is passed through a projection head to regress the 2-DoF offset
$\Delta_t=(\delta_x,\delta_y)\in\mathbb{R}^2$ within the retrieved tile.

\subsection{Two-stage Training and End-to-End Inference}\label{subsec:training}

\subsubsection{Two-stage Training}
We train TOLoc in two stages. In the first stage, only the PR module is optimized to retrieve the OSM tile that best matches each textual query, thereby establishing global text--map correspondences. In the second stage, the top-1 retrieved tile is fed into the PE module for fine-grained localization. The PR and PE modules are then jointly optimized, enabling the model to preserve reliable retrieval performance while learning to regress the final 2-DoF position.

\subsubsection{Loss Functions}
For place recognition, we use a symmetric cross-entropy loss over a mini-batch of $B$ paired samples $\{(\mathcal{T}_k,\mathcal{O}_k)\}_{k=1}^{B}$. Given the similarity matrix $S\in\mathbb{R}^{B\times B}$ computed by Eq.~(\ref{eq:cos_sim}), the text-to-map and map-to-text matching probabilities are
\begin{equation}
\begin{aligned}
P_{\mathcal{T}_k \rightarrow \mathcal{O}_l}
&=
\frac{\exp(S_{kl}/\tau)}
{\sum_{l=1}^{B}\exp(S_{kl}/\tau)},\\
P_{\mathcal{O}_l \rightarrow \mathcal{T}_k}
&=
\frac{\exp(S_{lk}/\tau)}
{\sum_{k=1}^{B}\exp(S_{lk}/\tau)},
\end{aligned}
\end{equation}
where $\tau$ is a temperature parameter. The PR loss is defined as
\begin{equation}
\begin{aligned}
\mathcal{L}_{\mathrm{PR}}
&=
\mathcal{L}_{\mathrm{t}\rightarrow\mathrm{m}}
+
\mathcal{L}_{\mathrm{m}\rightarrow\mathrm{t}}, \\
\mathcal{L}_{\mathrm{t}\rightarrow\mathrm{m}}
&= 
-\frac{1}{B}\sum_{j=1}^{B}
\log P_{\mathcal{T}_j \rightarrow \mathcal{O}_j},\\
\mathcal{L}_{\mathrm{m}\rightarrow\mathrm{t}}
&=
-\frac{1}{B}\sum_{k=1}^{B}
\log P_{\mathcal{O}_k \rightarrow \mathcal{T}_k}.
\end{aligned}
\end{equation}

For pose estimation, the model predicts a 2-DoF offset $\Delta_t=(\delta_x,\delta_y)$ relative to the center of the top-1 retrieved tile $\boldsymbol{\xi}_{\mathcal{O}}^{\text{top-1}}=(x_{\mathcal{O}},y_{\mathcal{O}})$. The final global position is
\begin{equation}\label{eq:infer}
\hat{\boldsymbol{\xi}}
=
\boldsymbol{\xi}_{\mathcal{O}}^{\text{top-1}}+\Delta_t.
\end{equation}
The PE module is supervised by an $L_1$ loss:
\begin{equation}
\mathcal{L}_{\mathrm{PE}}
=
\frac{1}{\sum_{k=1}^{B}\mathbb{I}_k}
\sum_{k=1}^{B}
\mathbb{I}_k
\left\|
\hat{\boldsymbol{\xi}}_k-\boldsymbol{\xi}_k
\right\|_1,
\end{equation}
where $\hat{\boldsymbol{\xi}}_k$ and $\boldsymbol{\xi}_k$ are the predicted and ground-truth positions of the $k$-th query. The validity mask $\mathbb{I}_k$ is defined as
\begin{equation}
\mathbb{I}_k=
\begin{cases}
1, & 
\left\|
\boldsymbol{\xi}_{\mathcal{O}_k}^{\text{top-1}}-\boldsymbol{\xi}_k
\right\|_1 < \epsilon,\\
0, & \text{otherwise},
\end{cases}
\end{equation}
where $\epsilon$ is a predefined threshold. Thus, only samples whose top-1 retrieved tile is close enough to the ground truth are used to optimize the PE module.

\subsubsection{End-to-End Inference}
At inference, the textual query is encoded online, while all OSM tile descriptors and patch-level features are precomputed. We retrieve the top-$K$ candidate tiles by computing text--map similarities using Eq.~(\ref{eq:cos_sim}). The TOA module then fuses the text descriptor with the patch-level features of the top-1 tile to predict the 2-DoF offset, from which the final global position is obtained by Eq.~(\ref{eq:infer}).

\section{Experiments}\label{sec:exp}

\begin{table}[t]
\footnotesize
\setlength{\tabcolsep}{3.pt}
\centering
\renewcommand{\arraystretch}{1.0}
\caption{Place recognition results on the TOL-N set.}
\label{tab:pr_results_n}
\begin{tabular*}{\columnwidth}{@{\extracolsep{\fill}}l c|ccc|ccc@{}}
\hline
\multirow{2}{*}{Method} & \multirow{2}{*}{Backbone} 
& \multicolumn{3}{c|}{$\eta=25$ m} 
& \multicolumn{3}{c}{$\eta=10$ m} \\ 
\cline{3-8}
& & R@1 & R@5 & R@10 & R@1 & R@5 & R@10 \\ \hline
GOTPR & GNN 
& 11.53 & 22.91 & 29.91 & 6.39 & 14.36 & 18.91 \\ \hline
\multirow{2}{*}{CT2Loc} 
& CLIP-B32 & 13.92 & 31.28 & 38.87 & 6.77 & 17.60 & 24.31 \\
& CLIP-B16 & 18.24 & 35.34 & 44.50 & 8.75 & 20.02 & 27.63 \\ \hline
\multirow{4}{*}{TOLoc}
& CLIP-B32 & 24.80 & 37.61 & 45.00 & 12.11 & 23.78 & 28.95 \\
& CLIP-B16 & 26.29 & 40.06 & 45.93 & 15.87 & 26.79 & 32.21 \\
& SigLIP-224 & 26.76 & \textbf{40.44} & 46.60 & 14.15 & 27.11 & 32.62 \\
& SigLIP-384 & \textbf{27.60} & 39.33 & \textbf{47.86} & \textbf{17.25} & \textbf{27.52} & \textbf{32.68} \\ \hline
\end{tabular*}
\vspace{-6pt}
\end{table}

\subsection{Experimental Setup}

All experiments are conducted on the TOL benchmark. We train all methods on the TOL-N training split, evaluate them on the TOL-N validation split, and further assess cross-scene generalization on TOL-K360.

We instantiate TOLoc with CLIP~\cite{radford2021learning} and SigLIP~\cite{zhai2023sigmoid} backbones. Specifically, we use \texttt{ViT-B/16} and \texttt{ViT-B/32} for CLIP, denoted as \texttt{TOLoc-C-B16} and \texttt{TOLoc-C-B32}, and \texttt{SigLIP-base-Patch16-224} and \texttt{SigLIP-base-Patch16-384} for SigLIP, denoted as \texttt{TOLoc-S-B224} and \texttt{TOLoc-S-B384}. All models are initialized from official pretrained weights and fine-tuned on the TOL training set. We train the PR module for 20 epochs, followed by joint optimization of the full model for another 20 epochs. We use Adam with a learning rate of $1\times10^{-5}$ and a batch size of 64. All experiments are conducted on two NVIDIA RTX 4090 GPUs.

\subsection{Evaluation of Place Recognition}

\subsubsection{Metrics}
For place recognition, we report Recall@K under a revisit threshold $\eta$. A query is considered correctly retrieved if at least one of the top-$K$ retrieved OSM tiles lies within $\eta$ meters of the ground-truth position. We report Recall@K for $K\in\{1,5,10\}$ under $\eta\in\{10,25\}$.

\subsubsection{Baselines}
We compare TOLoc with two recent T2O place recognition methods: GOTPR~\cite{jung2025gotpr} and CrossText2Loc from CVG-Text~\cite{ye2025cross}, denoted as CT2Loc. GOTPR performs scene-graph-based retrieval by modeling objects, attributes, and pairwise spatial relations between text and OSM maps. Since its scene graph construction code is unavailable, we reproduce the pipeline following the original paper and apply it to TOL under the observer-centric setting. CT2Loc is a contrastive learning framework for text-to-satellite/OSM retrieval; we concatenate the multiple hint sentences of each query into a single paragraph as input. For fair comparison, both baselines are trained using their official implementations with a batch size of 64.

\subsubsection{Results}
As shown in Tab.~\ref{tab:pr_results_n}, TOLoc consistently outperforms existing T2O place recognition methods. Compared with CT2Loc using the same CLIP backbone, TOLoc achieves higher recall, demonstrating the effectiveness of direction-aware text encoding and map feature aggregation. GOTPR performs relatively worse, likely due to the limited capacity of scene-graph matching in complex urban scenes.

Among CLIP variants, \texttt{CLIP-B16} outperforms \texttt{CLIP-B32}, suggesting that smaller patch sizes better preserve fine-grained spatial details for local layout reasoning. A similar trend appears for SigLIP, where \texttt{SigLIP-384} performs better than \texttt{SigLIP-224}. Overall, SigLIP-based variants achieve the best results, indicating the benefit of stronger pretrained representations.

\subsubsection{Generalization}
Results in Tab.~\ref{tab:gen_pr_n} show that both our method and the baselines generalize reasonably well to unseen regions, indicating that semantic and topological cues in T2O place recognition transfer across cities. Notably, models trained on limited geographic regions still achieve nontrivial performance in previously unseen cities that are geographically distant from the training set. Across all test sequences, TOLoc consistently outperforms the baselines, demonstrating superior accuracy and generalization.

\begin{table*}[t]
\caption{Cross-scene generalization of place recognition on the TOL-K360 set, measured by R@1 under the 25 m threshold.}
\centering
\renewcommand{\arraystretch}{1}
\footnotesize
\label{tab:gen_pr_n}
“00” denotes the 00 sequence of the TOL-K360 set, and the other sequences follow the same naming convention.

\begin{tabular*}{\textwidth}{@{\extracolsep{\fill}}lcccccccccccc@{}}
\hline
Method & Backbone & 00 & 02 & 03 & 04 & 05 & 06 & 07 & 08 & 09 & 10 & 18 \\ \hline
GOTPR & GNN & 2.11 & 4.16 & 30.20 & 4.94 & 9.28 & 5.21 & 12.91 & 7.04 & 4.33 & 11.27 & 7.60 \\ \hline

\multirow{2}{*}{CT2Loc}
& CLIP-B32 & 4.54 & 5.64 & 30.79 & 5.75 & 11.60 & 7.62 & 9.41 & 10.56 & 6.33 & 12.79 & 13.19 \\
& CLIP-B16 & 6.21 & 5.46 & 32.28 & 6.81 & 12.00 & 7.52 & 9.00 & 12.66 & 10.77 & 15.43 & 12.87 \\ \hline

\multirow{4}{*}{TOLoc}
& CLIP-B32 & 6.22 & 7.86 & 31.98 & 9.44 & 13.43 & 10.53 & 14.01 & 12.11 & 11.07 & 21.81 & 19.44 \\
& CLIP-B16 & 10.12 & 9.84 & 36.63 & 10.22 & 13.81 & 11.28 & 16.99 & 17.26 & 11.91 & 23.93 & 16.67 \\
& SigLIP-224 & \textbf{13.26} & 9.83 & 35.35 & 10.41 & 14.37 & 11.04 & 16.44 & 16.19 & 13.92 & \textbf{26.07} & \textbf{20.01} \\
& SigLIP-384 & 9.81 & \textbf{11.15} & \textbf{40.69} & \textbf{12.72} & \textbf{15.29} & \textbf{11.64} & \textbf{17.47} & \textbf{17.68} & \textbf{14.19} & 25.88 & 19.55 \\ \hline
\end{tabular*}
\end{table*}

\subsection{Evaluation of Localization}

\subsubsection{Metrics}
For localization, the model predicts the 2-DoF position within the top-1 retrieved OSM tile. We evaluate localization performance using two metrics: success rate (SR) and localization error (LE). SR measures the percentage of queries whose localization error is smaller than a predefined threshold. We report SR under thresholds of {5, 10, 25} meters. LE is reported using the 5th, 10th, and 25th percentiles of the localization error distribution over the test set.

\subsubsection{Baselines}
Since no prior work explicitly addresses text-to-OSM pose estimation, we design two PE-stage baselines. \textbf{CLS} matches the text descriptor and map features token-wise to predict a spatial probability distribution from a $7\times7$ similarity heatmap for each tile in \texttt{TOLoc-C-B32}. \textbf{MLP} concatenates the text and map descriptors and regresses the 2-DoF position with an MLP.
For GOTPR and CT2Loc, which are originally designed for place recognition only, we use center coordinates of the top-1 retrieved OSM tile as the predicted position and report the resulting localization performance.

\begin{table}[!t]
\setlength{\tabcolsep}{2.3pt}
\renewcommand{\arraystretch}{1}
\footnotesize
\centering
\caption{Localization results on the TOL-N set.}
\label{tab:loc_result_n}
\begin{tabular*}{\columnwidth}{@{\extracolsep{\fill}}lc|ccc|ccc@{}}
\hline
\multirow{2}{*}{Method} & \multirow{2}{*}{PE} &  \multicolumn{3}{c|}{SR$\uparrow$}   & \multicolumn{3}{c}{LE$\downarrow$} \\ \cline{3-8} 
            &            & @5m       & @10m  & @25m  & $@5\%$    & $@10\%$   & $@25\%$  \\ \hline
GOTPR        &    \texttimes     & 2.42     & 6.39  & 11.53 &  8.42  &  17.97  & 178.03  \\ 
CT2Loc-B32        &    \texttimes     & 2.16     & 7.32 & 15.35 &  8.67  & 13.51   & 111.95  \\  
CT2Loc-B16        &    \texttimes     & 2.19     & 8.75  & 18.23  &  7.17   &  11.02   & 76.67   \\  
TOLoc-C-B32           &    \texttimes     & 3.74    & 12.11  & 24.80  &  5.99   & 8.62    &  26.16 \\  
TOLoc-C-B16        &    \texttimes     &  5.19    & 15.87  & 26.29  & 4.94    & 6.92     & 19.40   \\ 
TOLoc-S-B224        &    \texttimes     &  4.76    & 14.15  & 26.76  & 5.14    & 7.88     & 20.29   \\ 
TOLoc-S-B384        &    \texttimes     &  6.22    & 17.25  & 27.60  & 4.50    & 6.97     & 17.25   \\  \hline
\multirow{4}{*}{TOLoc-C-B32}          &    CLS     &  3.36     & 11.79  & 21.97 & 6.19  & 9.02   & 46.54   \\
           &    MLP    &  7.00   & 13.74 & 21.59  & 4.26   & 6.84   & 45.85         \\
           &   TOA    & 7.21     & 17.01 & 24.63 & 4.07    & 6.58   & 26.44  \\ \hline
TOLoc-C-B16 & TOA & \textbf{8.72} & \textbf{18.68} & 26.55 & \textbf{3.51} & \textbf{5.38} & 18.53 \\ 
TOLoc-S-B224 & TOA & 6.42 & 17.30 & 26.82 & 4.35 & 6.56 & 18.59\\ 
TOLoc-S-B384 & TOA  & 6.83 & 18.65 & \textbf{28.10} & 4.30 & 6.43 & \textbf{15.75} \\ \hline
\end{tabular*}
\vspace{-4pt}
\end{table}

\subsubsection{Results}
Tab.~\ref{tab:loc_result_n} summarizes the localization results of TOLoc and its variants. PR-only methods, including GOTPR, CT2Loc, and TOLoc without the PE module, show limited localization accuracy and relatively large localization errors, indicating that place recognition alone is insufficient for accurate meter-level localization. Among PE-equipped variants, \textbf{CLS} predicts a spatial probability distribution from fused features, but lacks explicit local feature alignment for distinguishing fine-grained positions within the retrieved tile. As a result, it tends to collapse to the tile center and performs poorly under strict 5 m and 10 m thresholds. This suggests that fine-grained localization requires explicit local alignment rather than direct heatmap prediction from fused descriptors. \textbf{MLP} achieves better results by regressing position offsets with a dedicated PE stage. The full TOLoc model with the proposed TOA module obtains the best overall performance, demonstrating the effectiveness of explicit text--map alignment for fine-grained localization.

\begin{table*}[t]
\caption{Cross-scene generalization of localization on the TOL-K360 set under the 10 m threshold.}
\label{tab:gen_loc_k}
\centering
\setlength{\tabcolsep}{4pt}
\renewcommand{\arraystretch}{1}
\footnotesize
\begin{tabular*}{\textwidth}{@{\extracolsep{\fill}}l c ccccccccccc@{}}
\hline
Method & PE & Seq 00 & Seq 02 & Seq 03 & Seq 04 & Seq 05 & Seq 06 & Seq 07 & Seq 08 & Seq 09 & Seq 10 & Seq 18 \\ \hline
GOTPR          & \multirow{7}{*}{\texttimes} 
& 0.88 & 2.57 & 15.45 & 2.26 & 4.12 & 2.41 & 7.13 & 2.77 & 1.54 & 4.60 & 2.51 \\
CT2Loc-C-B32   &  
& 1.32 & 2.86 & 12.87 & 2.15 & 4.80 & 3.35 & 3.74 & 4.70 & 2.12 & 5.09 & 4.15 \\
CT2Loc-C-B16   &  
& 2.33 & 2.51 & 15.25 & 2.67 & 4.59 & 3.15 & 3.36 & 5.40 & 3.40 & 5.22 & 4.12 \\
TOLoc-C-B32    &  
& 2.37 & 4.46 & 13.86 & 3.97 & 4.16 & 5.86 & 5.71 & 6.16 & 4.40 & 8.23 & 7.49 \\
TOLoc-C-B16    &  
& 4.20 & 5.96 & 18.91 & 4.86 & 6.18 & 7.01 & 7.96 & 9.70 & 4.64 & 9.48 & 6.13 \\
TOLoc-S-B224   &  
& 6.42 & 6.02 & 20.30 & 5.17 & 6.85 & 7.07 & 7.20 & 9.41 & 6.56 & 12.69 & 7.17 \\
TOLoc-S-B384   &  
& 3.99 & \textbf{9.08} & 22.38 & 6.55 & 8.09 & 6.95 & 7.22 & 10.30 & 6.21 & 11.34 & 7.95 \\ \hline
TOLoc-C-B32    & \multirow{4}{*}{\checkmark} 
& 2.63 & 5.04 & 19.41 & 5.36 & 7.30 & 6.32 & 8.03 & 8.26 & 6.61 & 12.76 & \textbf{12.09} \\
TOLoc-C-B16    &  
& 5.50 & 6.45 & 22.97 & 6.13 & 9.38 & \textbf{8.35} & 8.30 & 11.61 & 7.26 & 14.84 & 9.42 \\
TOLoc-S-B224   &  
& \textbf{7.41} & 6.87 & 20.50 & 6.31 & 8.50 & 7.82 & 8.20 & 11.06 & 9.19 & \textbf{16.09} & 11.46 \\
TOLoc-S-B384   &  
& 4.46 & 7.49 & \textbf{26.73} & \textbf{7.95} & \textbf{10.00} & 8.07 & \textbf{9.52} & \textbf{12.16} & \textbf{10.89} & 13.62 & 8.93 \\ \hline
\end{tabular*}
\end{table*}

\begin{figure*}[t]
\vspace{-6pt}
    \centering
    \includegraphics[width=0.9\linewidth]{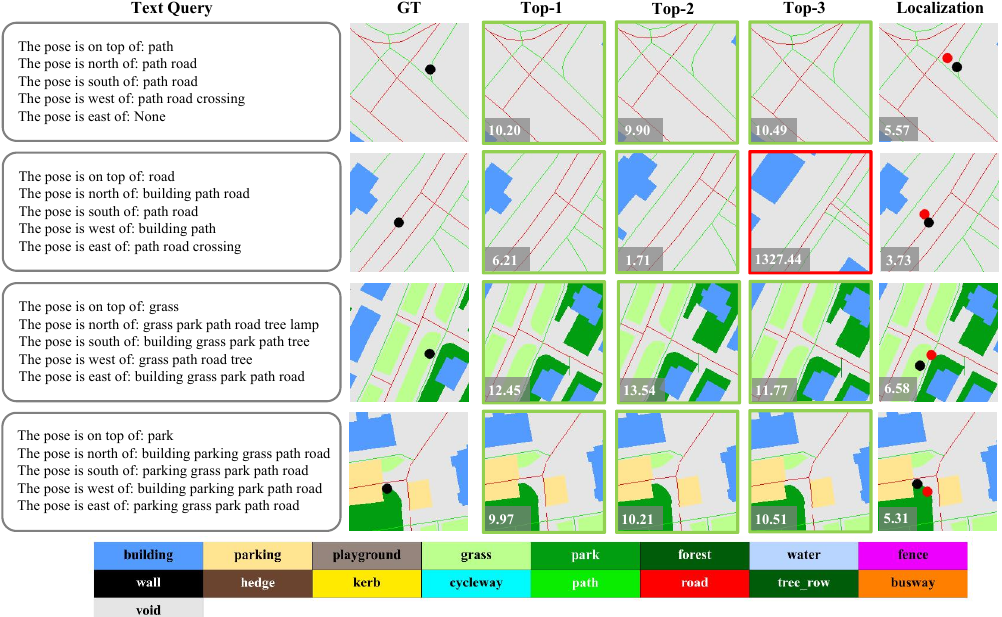}
  \caption{Qualitative results on the TOL benchmark.  In columns 3--5, green boxes \( \raisebox{0.5ex}{\textcolor{green}{\fbox{}}} \) indicate correct retrievals (error $\leq$ 25 m), while red boxes \( \raisebox{0.5ex}{\textcolor{red}{\fbox{}}} \) indicate failures. In columns 2 and 6, black dots $\textcolor{black}{\bullet}$ denote ground-truth query locations and red dots $\textcolor{red}{\bullet}$ denote the estimated positions. Retrieval and localization errors (in meters) are reported in the lower-left corner.}
    \label{fig:qua_loc}
\end{figure*}

\subsubsection{Generalization}
As shown in Tab.~\ref{tab:gen_loc_k}, all methods are trained on TOL-N and directly evaluated on TOL-K360 for cross-scene generalization. For PR-only methods, we use the center of the top-1 retrieved OSM tile as the final prediction. The results show that meter-level T2O localization in unseen scenes remains challenging. GOTPR achieves relatively competitive generalization, possibly because its scene graphs explicitly model map topology and semantic relations. Nevertheless, the full TOLoc model consistently outperforms all baselines, demonstrating the effectiveness of the proposed TOA module for fine-grained localization. Qualitative examples are shown in Fig.~\ref{fig:qua_loc}.

\subsection{Discussion}
We present ablation studies and failure case analysis on TOL-N to further evaluate the design choices. Unless otherwise specified, \texttt{TOLoc-C-B32} is used as the default variant, and Recall@$\{1,5,10\}$ is reported under the 25 m threshold.

\begin{table}[t]
\centering
\footnotesize

\centering
\setlength{\tabcolsep}{10pt}
\renewcommand{\arraystretch}{1}
\caption{Ablation study on text offset range.}
\label{tab:offset}
\begin{tabular}{l|ccc}
\hline
Text offset & R@1 & R@5 & R@10 \\ \hline
$1/12S$ & 53.55 & 73.33 & 82.29 \\
$1/6S$  & 40.53 & 60.23 & 70.06 \\
$1/3S$  & 28.83 & 49.75 & 59.94 \\
$1/2S$  & 21.59 & 36.10 & 43.92 \\ \hline
\end{tabular}

\vspace{10pt}

\centering
\setlength{\tabcolsep}{10pt}
\renewcommand{\arraystretch}{1}
\caption{Ablation study on text fusion order.}
\label{tab:order}
\begin{tabular}{l|ccc}
\hline
Text order & R@1 & R@5 & R@10 \\ \hline
NESWT & 21.45 & 33.91 & 42.69 \\
TNESW & 20.72 & \textbf{36.77} & 43.82 \\
TNWSE & 18.12 & 35.80 & 42.82 \\
TNSWE & \textbf{21.59} & 36.10 & \textbf{43.92} \\ \hline
\end{tabular}

\vspace{10pt}

\centering
\setlength{\tabcolsep}{8pt}
\renewcommand{\arraystretch}{1}
\caption{Runtime performance analysis.}
\label{tab:runtime}
\begin{tabular}{ccccc}
\hline
\multicolumn{2}{c}{Module} & Param. & Com. & Run. \\ \hline
\multicolumn{1}{c|}{\multirow{2}{*}{PR}} & Text encoder & 64.74M & 14.52G & 52.35 \\
\multicolumn{1}{c|}{} & Map encoder & 91.39M & 4.54G & 18.01 \\ \hline
\multicolumn{2}{c}{PE} & 19.94M & 0.36G & 2.80 \\ \hline
\multicolumn{2}{c}{Total} & 176.07M & 19.42G & 73.16 \\ \hline
\end{tabular}
\vspace{-4pt}
\end{table}

\subsubsection{Text Offset}
The text-query position may deviate from the center of the ground-truth OSM tile, causing text--map inconsistency. We study this effect by varying the query-to-center offset and report the results in Tab.~\ref{tab:offset}. Larger offsets reduce retrieval accuracy, indicating that stronger spatial inconsistency makes text--map matching more challenging.

\subsubsection{Text Fusion Order}
The final text descriptor is obtained by concatenating directional text features, whose order may affect the representation. We evaluate different ordering strategies in Tab.~\ref{tab:order}. The results show only minor performance variations across permutations, suggesting that the model is not highly sensitive to the fusion order. We adopt \texttt{TNSWE}, which achieves the highest top-1 recall.

\subsubsection{Runtime Performance}
Tab.~\ref{tab:runtime} reports the parameters, FLOPs, and runtime of \texttt{TOLoc-C-B32}, measured on an NVIDIA RTX 4090 GPU and an Intel i9-13900K CPU using \texttt{fvcore}. Most parameters and computation lie in the PR module, which encodes text and OSM tiles into descriptors. A single text--map descriptor construction takes about 70 ms, while the PE module is lightweight and requires only 2.80 ms for feature fusion and offset prediction.

\section{Conclusion}\label{sec:conclusion}
In this work, we formulate the text-to-OSM localization task and introduce TOL, a large-scale benchmark specifically designed for this problem. TOL provides approximately 121K textual descriptions paired with OSM map data across diverse regions, enabling systematic evaluation of text-based localization. To address this task, we further propose TOLoc, a two-stage localization framework that follows a coarse-to-fine paradigm, combining text–map place recognition with fine-grained pose estimation through cross-modal alignment.
Experimental results show that TOLoc achieves accurate localization and strong generalization, outperforming baseline methods by a large margin. We hope this work will facilitate future research on scalable text-based localization.

\bibliographystyle{IEEEtran}

\bibliography{glorified,tol}

\end{document}